\documentclass{article}

\usepackage[preprint]{neurips_2020}

\usepackage[ruled,vlined]{algorithm2e}
\usepackage[utf8]{inputenc} % allow utf-8 input
\usepackage[T1]{fontenc}    % use 8-bit T1 fonts
\usepackage{hyperref}       % hyperlinks
\usepackage{url}            % simple URL typesetting
\usepackage{booktabs}       % professional-quality tables
\usepackage{amsfonts}       % blackboard math symbols
\usepackage{amsmath}
\usepackage{nicefrac}       % compact symbols for 1/2, etc.
\usepackage{microtype}      % microtypography
\usepackage{siunitx}
\usepackage{natbib}
\usepackage{graphicx}
\usepackage{multirow}

\DeclareMathOperator*{\argmax}{arg\,max}

\title{Representation Learning with Video Deep InfoMax}

\author{%
  R Devon Hjelm\\
  Microsoft Research\\
  \And
  Philip Bachman \\
  Microsoft Research \\
}

\begin{document}

\maketitle

\begin{abstract}
    Self-supervised learning has made unsupervised pretraining relevant again for difficult computer vision tasks. The most effective self-supervised methods involve prediction tasks based on features extracted from diverse views of the data. Deep InfoMax (DIM) is a self-supervised method which leverages the internal structure of deep networks to construct such views, forming prediction tasks between local features which depend on small patches in an image and global features which depend on the whole image. In this paper, we extend DIM to the video domain by leveraging similar structure in spatio-temporal networks, producing a method we call \emph{Video Deep InfoMax} (VDIM). We find that drawing views from both natural-rate sequences and temporally-downsampled sequences yields results on Kinetics-pretrained action recognition tasks which match or outperform prior state-of-the-art methods that use more costly large-time-scale transformer models. We also examine the effects of data augmentation and fine-tuning methods, accomplishing SoTA by a large margin when training only on the UCF-$101$ dataset.
\end{abstract}

\section{Introduction}
We begin by considering difficult real-world machine learning tasks involving high-dimensional vision data~\citep[e.g.,][to name a few]{soomro2012ucf101, lin2014microsoft, Russakovsky2015, Everingham15, krishna2017visual, song2016ssc, sharma2018conceptual, sun2019scalability}, such as those used to train complex vision systems to make decisions in real-world environments.
%One of the key properties of successful vision systems is they have internal representations that are useful, in the sense that information encoded in these representations easily translates to correct decisions during important downstream tasks.
Successful vision systems often rely on internal representations that encode information that translates, through further processing, to correct decisions on these downstream tasks.
Training systems to encode useful information is a core focus of research on \emph{representation learning}, which has provided crucial support for the surge in useful AI tools across numerous domains~\citep{brock2018large, tan2019efficientnet, liu2019roberta, clark2019adversarial, brown2020language}.

While supervised learning has received significant attention, unsupervised learning holds great promise for representation learners because it derives the learning signal from implicit structure in the data itself, rather than from external human annotation~\citep{goodfellow2016deep}. 
Annotation can be problematic for various reasons: because it is expensive to collect~\citep[][]{zhu2009introduction}, because it is prone to human error~\citep[][]{devillers2005challenges, nettleton2010study}, or because it provides no guarantees when transferring the resulting representations to other tasks~\citep{sylvain2019locality}.

Unsupervised learning helps to discover representations that more closely resemble a code of the underlying factors of the data, thus implying a broader range of utility in deployment scenarios~\citep{bengio2013representation}.
For instance, successfully identifying an airplane in an image could rely on identifying sky in the background -- a feature that may not translate well to other tasks.
In contrast, models capable of generating data representative of
a full target distribution must contain some understanding of the underlying factors, such as the parts of the airplane and their spatial relationships.
Since this kind of understanding encompasses simpler tasks like classification, we expect models that encode these factors to be able to classify well~\citep{donahue2016adversarial, dumoulin2016adversarially, donahue2019large} as a corollary, and to perform well on other downstream tasks~\citep{ulusoy2005generative, pathak2016context, radford2018improving}.

Self-supervised learning~\citep{doersch2015unsupervised, doersch2017multi, kolesnikov2019revisiting, goyal2019scaling} involves unsupervised methods which procedurally generate supervised tasks based on informative structure in the underlying data.
These methods often require generating multiple \emph{views} from the data, e.g.~by: selecting random crops~\citep{bachman2019learning}, shuffling patches~\citep{noroozi2016unsupervised}, rotating images~\citep{jing2018self}, removing color information~\citep{zhang2016colorful}, applying data augmentation~\citep{tian2019contrastive}, etc. 
The model is then trained to predict various relationships among these views, such as identifying whether two random crops belong to the same image, predicting how to order shuffled patches, predicting the way an image has been rotated, etc.
Deciding how to generate useful views is central to the success of self-supervised learning, as not all view-based tasks will encourage the model to learn good features -- e.g., models can ``cheat" on colorization tasks in the RGB space~\citep{zhang2016colorful}.
Despite this challenge, self-supervised learning has led to accelerated progress in representation learning, making unsupervised methods competitive with supervised learning on many vision tasks~\citep{chen2020simple, caron2020unsupervised}.

While many of the success stories in self-supervised learning for vision come from research on static images, this setting seems inherently limited: static images contain little information about the dynamic nature of objects, their permanence and coherence, and their affordances compared to videos. 
%We therefore expect representations learned from video to capture distinct and richer information. (DEVON: unfortunately, we don't have much to say about this, so I'd rather not talk too much about it yet. I mention this briefly a few times, which is maybe even too much)
In this work, we explore learning video representations using Deep InfoMax~\citep{hjelm2018learning, bachman2019learning}, a contrastive self-supervised method that generates views by data augmentation and by selecting localized features expressed within a deep neural network encoder.
DIM provides an opportunity to explore the role of views in the spatio-temporal setting, as these can be generated by leveraging the natural structure available from a spatio-temporal convolutional neural network, in addition to data augmentation and subsampling techniques.
We use views in conjunction with a contrastive loss~\citep{oord2018representation} to learn representations in an unsupervised way, and we evaluate learned representations by fine-tuning them for downstream action classification.
We achieve comparable or better performance on action recognition with less pretraining compared to larger transformer models on the Kinetics $600$ dataset~\citep{carreira2018short}, and surpass prior work by a large margin with a model trained using only the UCF-$101$ dataset~\citep{soomro2012ucf101}.
We also provide a detailed ablation on the fine-tuning procedure, which is crucial to downstream performance but understudied in prior related work~\citep{han2019video, sun2019contrastive}.

\section{Background and Related Work}
Machine learning classification systems in vision often surpass human performance on static images~\citep{he2016deep, touvron2020fixing}, yet harder tasks that rely on understanding the objectness or affordances of an image subject or subjects, such as localization, semantic segmentation, etc. still pose major challenges for vision systems~\citep{Everingham15, hall2020probability}.
This is in contrast to human perception, where understanding these properties is at least as fundamental, if not more so, as the ability to reason about an object's class~\citep{tenenbaum2011grow}.
But, unlike machine learning systems, human experience of the world is dynamic and interactive, so it is perhaps unreasonable to expect a model to learn about objects and their properties without being exposed to data or interacting with it in a similar way. % (DEVON): I'd like to find a good paper for this statement, but most cognitive scientist argue against it. I disagree with them.

%For example, images of airplanes or boats commonly appear with a background of sky or sea, respectively, but without strong priors (such as encouraging the model to focus on central pixels), there's nothing to guide the model on focusing on important things, like objects, while paying less attention to less important things, like background.
%Annotation and supervised learning provides a strong prior, but this relies a great deal on knowing which annotations are useful and how this translates to deployment scenarios.
%However, if one is able to observe an airplane from many different ``views" over time, such as during boarding and fueling, taking off, flying, landing, and deplaning, this surly admits a more unsupervised approach to learn representations that encode a more comprehensive understanding of these objects.

One wonders then why so many important vision systems meant to function in a dynamic world are largely based on models trained on static images~\citep{yuan2019object}.
The reasons have been mostly practical and computational.
%If it weren't for the obvious computational and other practical challenges.
One obvious challenge is that annotation for dynamic/interactive data, like video, is more time-consuming and costly to collect~\citep{vondrick2013efficiently}.
%as the addition of time and the necessary multi-object nature of video adds additional challenges for human annotators beyond the normal challenges of annotating static image data.
In addition, labels of dynamic processes recorded in video, such as action, intent, etc., can be highly subjective, and their annotation requires models to adhere to a particular human interpretation of a complex scene~\citep{reidsma2008exploiting}.
These challenges suggest that minimizing reliance on supervised learning is an appealing approach for learning %useful, dynamic
representation-based systems of video.

In this direction, unsupervised video representations have begun to show promise~\citep{wang2015unsupervised, srivastava2015unsupervised}.
Generative models for video have improved in recent work~\citep{clark2019adversarial}, but these may be too demanding: the ability to generate a dynamic scene with pixel-level detail may not be necessary for learning useful video representations.
\emph{Self-supervised learning} asks less of models and has made progress in learning useful video representations that perform well on a number of important benchmark tasks~\citep{zou2012deep, misra2016shuffle, fernando2017self, sermanet2018time, vondrick2018tracking, jing2018self, ahsan2019video, dwibedi2019temporal, el2019skip, kim2019self, xu2019self}.
Contrastive learning is a popular type of self-supervised learning, which has proven effective for video~\citep{gordon2020watching, knights2020temporally} and reinforcement learning tasks~\citep{oord2018representation, anand2019unsupervised, srinivas2020curl, mazoure2020deep, schwarzer2020data}.
The most successful contrastive methods for video borrow components from models that process language data, such as recurrent neural networks~\citep{han2019video} and transformers~\citep{sun2019contrastive}.
The aptness of these architectures and their associated priors for video is based on analogy to discrete, language-like data, but it is still unclear whether these models justify their expense in the video setting.
In contrast, we focus on contrastive learning using only convolutions to process the spatio-temporal data, leveraging the natural internal structure of a spatio-temporal CNN to yield diverse views of the data.
Finally, we use downsampling to generate multiple views as additional augmentation~\citep[reminiscent of slow-fast networks for video processing, see][]{feichtenhofer2019slowfast}, and we show that this is important to downstream performance.

\section{Method}
\begin{figure*}[ht]
\begin{minipage}{0.5\textwidth}
\includegraphics[width=0.98\columnwidth]{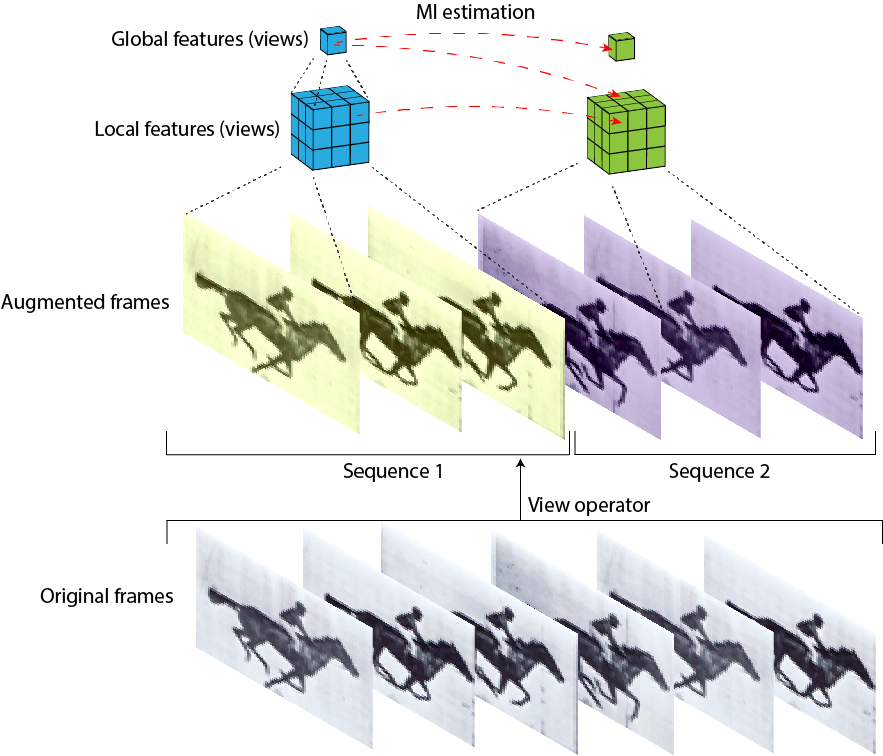}
\end{minipage}
\hfill\vline\hfill
\begin{minipage}{0.48\textwidth}
\includegraphics[width=0.95\columnwidth]{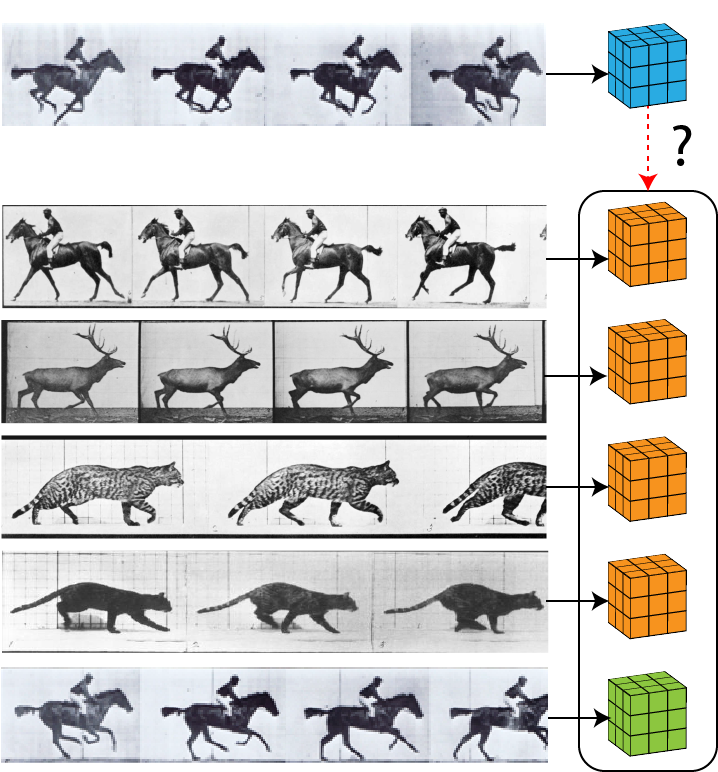}
\end{minipage}

\begin{minipage}{1.0\textwidth}
\caption{{\bf Left: High-level view of VDIM.} We first transform a sequence of frames from a video clip using a randomly generated view operator.
This splits the sequence into multiple subsequences, which may overlap, and applies random data augmentation to each subsequence.
We then process each subsequence independently using a spatio-temporal encoder to extract features representing different views of the data, which differ due to the random view operator and limited receptive fields in the encoder.
We then maximize mutual information between the features from the various views.
{\bf Right: Contrastive learning on local features.} Each set of local features are used in a set of k-way classification tasks. Each pair of local features from the ``antecedent" features (in blue, can be thought of as a ``query") and the correct ``consequent" features (in green, can be though of as the correct ``key") are compared against \emph{all} local features coming from negative samples.}
\label{fig:VDIM_model}
\end{minipage}
\end{figure*}

Our model draws components from Augmented Multiscale Deep InfoMax~\citep[AMDIM,][]{bachman2019learning}, whose architecture was specifically designed to encode features that depend on limited receptive fields.
As is standard in AMDIM, our model has three primary components: a ``view generator", a joint encoder + mutual information-estimating architecture, and a loss based on a lower bound on the mutual information. We illustrate our model in Figure~\ref{fig:VDIM_model}.

\subsection{The view generator}
We process the input, which in our case is a sequence of frames (each example is a 4D tensor with dimensions corresponding to time, height, width, and color channels), by a random augmentation operator that generates different views. This can be understood as a single operator $G: \mathcal{X} \to \prod \tilde{\mathcal{X}}$, which maps an input sequence in $\mathcal{X}$ to a set of subsequences in $\tilde{\mathcal{X}}$. For example, the operator might first sample two subsequences, apply downsampling to one or both, then apply pixel-wise data augmentation to every frame in each subsequence.

As is evident in other self-supervised and contrastive methods, the exact form of the ``view operator" $G$ is crucial to success in downstream tasks~\citep{tian2019contrastive, chen2020simple}. 
In this work, we compose $G$ from the following operations:
\begin{enumerate}
    \item First, a sequence of frames (a video clip) is drawn from the data. In order to standardize the input to a fixed number of time steps, we randomly draw a set of clips from a fixed length moving window with stride of 1. If the total clip length is less than this window size, we fill by repeating the last frame.
    \item The clip is then split into two possibly overlapping sequences of frames. The size of these subsequences depends on the downsampling we describe below, so these subsequences are chosen to ensure that each final subsequence is the same length. For example, for two disjoint subsequences, if the first subsequence is not downsampled and the second is downsampled by a factor of $2$, the first subsequence receives $1/3$ of the frames and the second receives the remaining $2/3$.
    \item Each subsequence is independently downsampled by a predetermined factor, resulting in two subsequences of equal length.
    \item Data augmentation is applied on each subsequence independently, with every frame within a subsequence being augmented with the same augmentation parameters. The augmentations chosen here are the standard ones from image tasks: random resized crop, random color jitter, random grey, and random rotation.
    As an alternative to random color jitter and grey, we can convert each frame from RGB to Lab, and then randomly drop a color channel.
\end{enumerate}
As the above augmentation is random, we can consider this process as drawing a view operator from a distribution (which we call the ``view generator"), such that $G \sim p(G)$. The view generator is specified by the operations, their ordering, as well as any parameters associated with random sampling, such as uniform sampling over brightness or contrast.
Because $G$ generates multiple views from the same input, we refer to the set of augmented views given input $x$ as $\{G_k(x)\} = G(x)$.

\subsection{Architecture}
Our architecture is a combination of those found in AMDIM and R(2+1)D~\citep{tran2018closer}, similar to S3D~\citep{xie2017rethinking} as used in our strongest baseline~\citep{sun2019contrastive}. R(2+1)D is a convolutional recurrent block that first applies 2D convolutions to the spatial dimension of feature maps, followed by a 1D convolution to the temporal dimension.
In our version, we take the AMDIM base architecture, expanding $1\times1$ convolutions that directly follow spatial convolutions to operate as a 1D convolutional layer in the temporal dimension.
This is more lightweight than using full 3D convolutions and maintains the property of the AMDIM architecture that features relatively deep in the network are still ``local" -- that is, they correspond to limited spatio-temporal patches of the input~\citep[i.e., do not have receptive fields that cover the complete input, as is the case with a standard ResNet, see][]{he2016deep}.

Let $\Psi: \tilde{\mathcal{X}} \to \prod_j \mathcal{F}_j$ denote our encoder, which transforms augmented video subsequences into a set of feature spaces, $\mathcal{F}_j$.
Each feature space is a product of \emph{local feature spaces}, $\mathcal{F}_j = \prod_{i=1}^{N_j} \mathcal{L}_{i, j}$, where the $\mathcal{L}_{i,j}$ are spanned by features that depend on a specific spatio-temporal ``patch" of the input.
The feature spaces are naturally ordered, such that features in later $\mathcal{F}_j$ become progressively less ``local" (they depend on progressively larger spatio-temporal patches) as the input is processed layer-by-layer through the encoder.
The number of local spaces also decreases (i.e., $N_j <= N_{j-1}; \forall j > 1$), finally resulting in a ``global" feature space that depends on the full input.
%The \emph{local features}, $f_{i, j} \in \mathcal{L}_{i, j}$ represent high-level encodings of the input corresponding to an augmented spatiotemporal ``patch" of the input.

\subsection{Mutual information estimation and maximization}
Mutual information can be estimated with a classifier~\citep{belghazi2018mutual, hjelm2018learning, oord2018representation, poole2019variational}, in our case by training a model to discern sets of features extracted from the same video clip from features drawn at random from the product of marginals over a set of clips.
Let us denote the local features independently processed by the encoder, $\Psi$, from the two augmented versions of an input sequence, $x \sim \mathcal{D}$, as $f_{i, j}(G_1(x))$ and $f_{i', j'}(G_2(x))$, resulting from applying $G \sim p(G)$ on $x$.
Let $\mathcal{J}$ denote a subset of all layers in the encoder where we will apply a mutual information maximization objective via a \emph{contrastive loss}.
We introduce a set of \emph{contrastive models}, each of which operate on a single layer across all locations, $\Phi = \{\phi_j: \mathcal{L}_{i, j} \to \mathcal{Y} | j \in \mathcal{J}\}$ meant to process the features to a space where we can compute pairwise scores (in our case, using a dot-product). As in AMDIM, $\Phi$ is modeled by MLPs with residual connections \footnote{We implement the location-wise MLP using $1 \times 1$ convolutions for efficient parallel processing.}.

We then estimate the mutual information between all pairs of local features from select layers: for every layer pair and location pair tuple, $(i, i', j, j')$, we compute a scalar-valued score as:
\begin{align}
    s^+_{i, i', j, j'}(x) = \phi_j(f_{i, j}(G_1(x))) \cdot \phi_{j'}(f_{i', j'}(G_2(x))).
    \label{eq:poseq}
\end{align}
Next, we require scores corresponding to ``negative" counterfactuals to be compared against the ``positive" pairs.
We compute these negative scores from pairs $(f_{i, j}(G_1(x), f_{i', j'}(G'_2(x'))$ obtained from two possibly different inputs, $x, x'$ (e.g., different subsequences from different video clips):
\begin{align}
   s^-_{i, i', j, j'}(x, x') = \phi_j(f_{i, j}(G_1(x)) \cdot \phi_j(f_{i', j'}(G'_2(x')),
   \label{eq:negeq}
\end{align}
where the view operator $G' \sim p(G)$ is drawn independently of $G$.
Let $X^{\prime}: x^{\prime} \sim \mathcal{D}$ denote a set of sequences sampled at random from the dataset.
The infoNCE lower-bound~\citep{oord2018representation} to the mutual information associated with the tuple $(i, i, j, j')$ is:
\begin{align}
    \Hat{\mathcal{I}}_{i, i', j, j'}(\mathcal{D}) = \mathbb{E}_{x \sim \mathcal{D}} \bigg[ \log \frac{\exp(s^+_{i, i', j, j'}(x))}{ \sum_{x' \in X^{\prime} \bigcup \{x\}} \sum_{i, i'} \exp(s^-_{i, i', j, j'}(x, x'))} \bigg]
\end{align}
Our final objective maximizes the mutual information across all locations over the complete set of ``antecedent" feature layers, $\mathcal{J}$, and ``consequent" feature layers, $\mathcal{J}'$:
\begin{align}
    \Psi^{\star}, \Phi^{\star} = \argmax_{\Psi, \Phi} \sum_{j \in \mathcal{J}, j' \in \mathcal{J}'} \sum_{i=1}^{N_j} \sum_{i'=1}^{N_{j'}} \Hat{\mathcal{I}}_{i, i', j, j'}(\mathcal{D}).
\end{align}
As our model is based on Deep InfoMax, we call this extension to spatio-temporal video data \emph{Video Deep InfoMax} (VDIM).
We draw the contrasting set $X^{\prime}$ from other samples in the same batch as $x$.

\section{Experiments}
We now demonstrate that models pretrained trained with Video Deep InfoMax (VDIM) perform well on downstream action recognition tasks for video data.
We first conduct an ablation study on the effect of the view generator, $p(G)$, showing that the type of augmentation has an effect on downstream performance.
We then pretrain on a large-scale video dataset, showing that such large-scale pretraining yields impressive results.

We use the following datasets in our study:
\begin{itemize}
    \item UCF-$101$~\citep{soomro2012ucf101} is an action recognition benchmarking dataset with $101$ classes and $13320$ videos. UCF-101 has three train-test splits. Videos are anywhere from a few seconds to minutes long (most are $5-10$ seconds). For Kinetics-pretrained models, we evaluate on all three splits, while for UCF-101 pretrained models, we pretrain and evaluate on only the first split to ensure that training samples are aligned across pretraining and downstream evaluation.
    \item HMDB-$51$~\citep[Human motion database,][]{kuehne2011hmdb} is another publicly available dataset for benchmarking action recognition models. HMDB is composed of $51$ human actions from $7000$ clips.
    HMDB-$51$ differs from UCF-$101$ in that the clips are generally much shorter (most are under $5$ seconds long).
    \item Kinetics $600$~\citep{carreira2018short} is a video dataset of roughly $600000$ videos from Youtube corresponding to $600$ action categories that is commonly used for pretraining and downstream evaluation. Unlike UCF-$101$ and HMDB-$51$, Kinetics is not a truly ``public" dataset, as access to all samples is difficult or impossible\footnote{For instance, when videos are removed from YouTube.}. Our version of Kinetics $600$ was acquired in March 2020, with $361080$ training videos and $54121$ test videos.
\end{itemize}

For all of the video datasets in training, we sample sequences of frames by first sampling a video clip, followed by subsampling a set of contiguous frames.
Each frame is resized to $128 \times 128$ during random-resized crop augmentation, and the final length of every sequence is chosen to ensure that subsequences that represent views are $32$ frames long after downsampling.
All models are optimized with Adam with a learning rate of \SI{2e-4}, and batch norm is used on the output of the encoder as well as in the contrastive models, $\phi_j$.

\begin{figure*}[ht]
\centering
\begin{minipage}{0.45\textwidth}
\centering
a) $\text{ConvBlock}(k, k', s, s', \text{res})$
\end{minipage}
\begin{minipage}{0.3\textwidth}
\centering
b) $\text{ResBlock}(k, k', s, s', n)$
\end{minipage}

\begin{minipage}{0.45\textwidth}
\centering
\includegraphics[width=0.7\columnwidth]{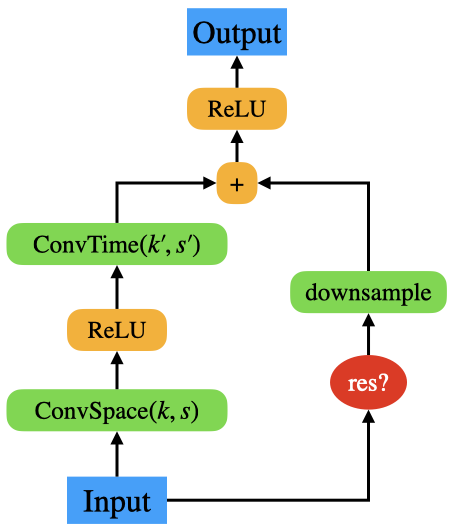}
\end{minipage}
\begin{minipage}{0.3\textwidth}
\centering
\includegraphics[width=0.7\columnwidth]{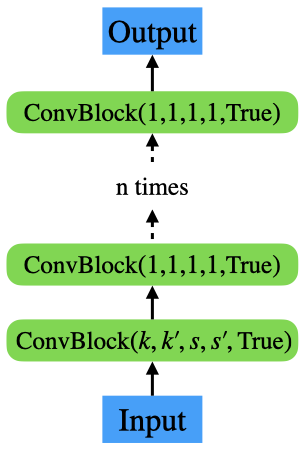}
\end{minipage}

\begin{minipage}{1.0\textwidth}
\caption{a) Basic convolutional block~\citep[R(2+1)D,][]{tran2018closer} used in our model. b) Residual block composed of a single residual convolutional block followed by $n$ $1\times1$ R(2+1)D blocks. The $1\times1$ convolutions help control the receptive fields of features, allowing VDIM to draw views by selecting high-level features.}
\label{fig:VDIM_model2}
\end{minipage}
\end{figure*}

\begin{table}[ht]
\centering
\caption{The full encoder model using components in Figure~\ref{fig:VDIM_model2}. $B\times T\times X \times Y$ are sizes in the batch, time, width, and height dimensions, respectively.}
\begin{tabular}{@{}|l|l|l|l|@{}}
\hline
{\bf Index}  & {\bf Block}                          & {\bf Size $(B \times T \times X \times Y)$}                                & {\bf View?}  \\
\hline
\hline
1 & ConvBlock$(5,5,2,2,\text{False})$     & $64 \times 16 \times 64 \times 64$  &        \\
2 & ConvBlock$(3,3,1,1,\text{False})$     & $64 \times 14 \times 62 \times 62$  &        \\
3 & ResBlock$(4,1,1,1,3)$          & $128 \times 14 \times 30 \times 30$ &        \\
4 & ResBlock$(4,1,1,1,3)$          & $256 \times 14 \times 14 \times 14$ &        \\
5 & ResBlock$(2,1,2,1,3)$          & $512 \times 7 \times 7 \times 7$    & Local 1  \\
6 & ResBlock$(3,1,3,1,3)$          & $512 \times 5 \times 5 \times 5$    & Local 2 \\
7 & ResBlock$(3,1,3,1,3)$          & $512 \times 3 \times 3 \times 3$    &        \\
8 & ResBlock$(3,1,3,1,0)$ & $512 \times 1 \times 1 \times 1$    & Global \\
\hline
\end{tabular}
\label{tab:model}
\end{table}

\begin{figure*}[ht]
\centering
\begin{minipage}{0.49\textwidth}
\centering
\includegraphics[width=0.75\columnwidth]{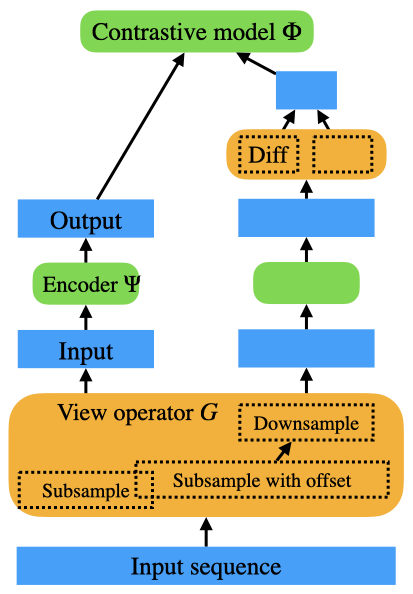}

\end{minipage}
\begin{minipage}{0.49\textwidth}
\centering
\includegraphics[width=0.95\columnwidth]{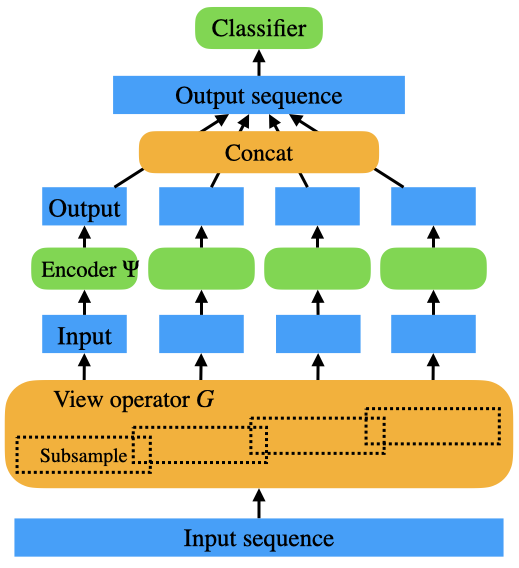}
\end{minipage}
\caption{{\bf Left: Pretraining} The view operator, $G \sim p(G)$ selects two subsequences, the second with a fixed offset, downsamples the second sequence, and applies data augmentation to both independently. The second subsequence is selected such that the final length after downsampling matches the first (in this case, for a final length of $32$ frames). These subsequences are used as input to the encoder, $\Psi$, and the outputs (from select antecedent feature layers $\mathcal{J}$ and consequent feature layers $\mathcal{J}'$) are used as input to a pair of contrastive models $\Phi$. In one variation of VDIM, the features from the second subsequence are split temporally, and their element-wise difference is sent to the contrastive model. {\bf Right: Classification} For downstream evaluation, $G$ selects a set of subsequences, each with a fixed offset, and downsampling is performed on each along with independent data augmenation. The output features of the encoder are concatenated (feature-wise) and used as input to a small classifier.}
\label{fig:pretrain}
\end{figure*}

Architecture details are provided in Table~\ref{tab:model}, with model components depicted in Figure~\ref{fig:VDIM_model2}.
The architecture is modeled after AMDIM, which limits features' receptive fields (we call these  local features) until the last two residual blocks (which we call global features, since their receptive fields cover the entire input).
We pick $\mathcal{J} = \{5, 6, 8\}$ to cover a reasonably representative set of views: one global and two locals with differently-sized receptive fields, while $\mathcal{J}' = \{8\}$ are chosen to correspond only to the global view to save memory.
The reverse, $\mathcal{J} = \{8\}$ and $\mathcal{J}' = \{5, 6, 8\}$, is also used.
The contrastive models, $\phi_j$, are residual MLPs with a single hidden layer, and the output features have size $|\mathcal{Y}| = 512$.

We present a high-level overview of the model used in pretraining in Figure~\ref{fig:pretrain} (left).
During pretraining, the view operator $G$ selects the first view as the first $32$ frames.
The second view is drawn from a fixed offset, either $0$ or $32$ frames.
The length of the second view is selected such that, after downsampling, the final length is $32$ frames.
The outputs of the first view are selected from $j \in \mathcal{J}$, the second are selected from $j' \in \mathcal{J}'$, and each contrastive model, $\phi_j, \phi_{j'}$ is used in Equations~\ref{eq:poseq} and \ref{eq:negeq} to yield the contrastive loss.
In one variant, the features of the second view are split in half temporally, and features constructed from the temporal difference are used in the contrastive loss.

\subsection{The effects of hyperparameters on fine-tuning}
On our set of benchmarks, key results presented in prior work come from first doing unsupervised pretraining with a large video dataset, then fine-tuning the encoder end-to-end on the benchmark task using a supervised loss.
We found that evaluation results varied widely depending on the fine-tuning procedure, mostly due to overfitting effects, because we lack any protections typically used in supervised learning such as dropout~\citep{srivastava2014dropout}.
Prior works rely on a decaying learning rate schedule~\citep{sun2019contrastive}, but do not specify the schedule and/or do not provide experimental results that support their schedules.
To fill this gap in understanding, we undertook an ablation study on relevant hyperparameters.

A high-level overview of downstream evaluation is given in Figure~\ref{fig:VDIM_model2} (right).
The view operator selects $K$ subsequences, each of which has an offset that is a multiple of $32$ (e.g., $0$, $32$, $64$, etc).
The subsequence is selected such that, after downsampling, the final length of each subsequence is $32$ frames.
Each subsequence is processed by the encoder, the outputs are concatenated, and the final set of features are sent to a classifier with a single hidden layer of size $1024$ with dropout ($p=0.2$).
For the final encoder at evaluation, we use $\phi_8 \circ \Psi$.
For action recognition evaluation on the test split, we average the class assignment probabilities across all samples from the same video clip.

\begin{figure*}[ht]
\centering
\includegraphics[width=0.75\columnwidth]{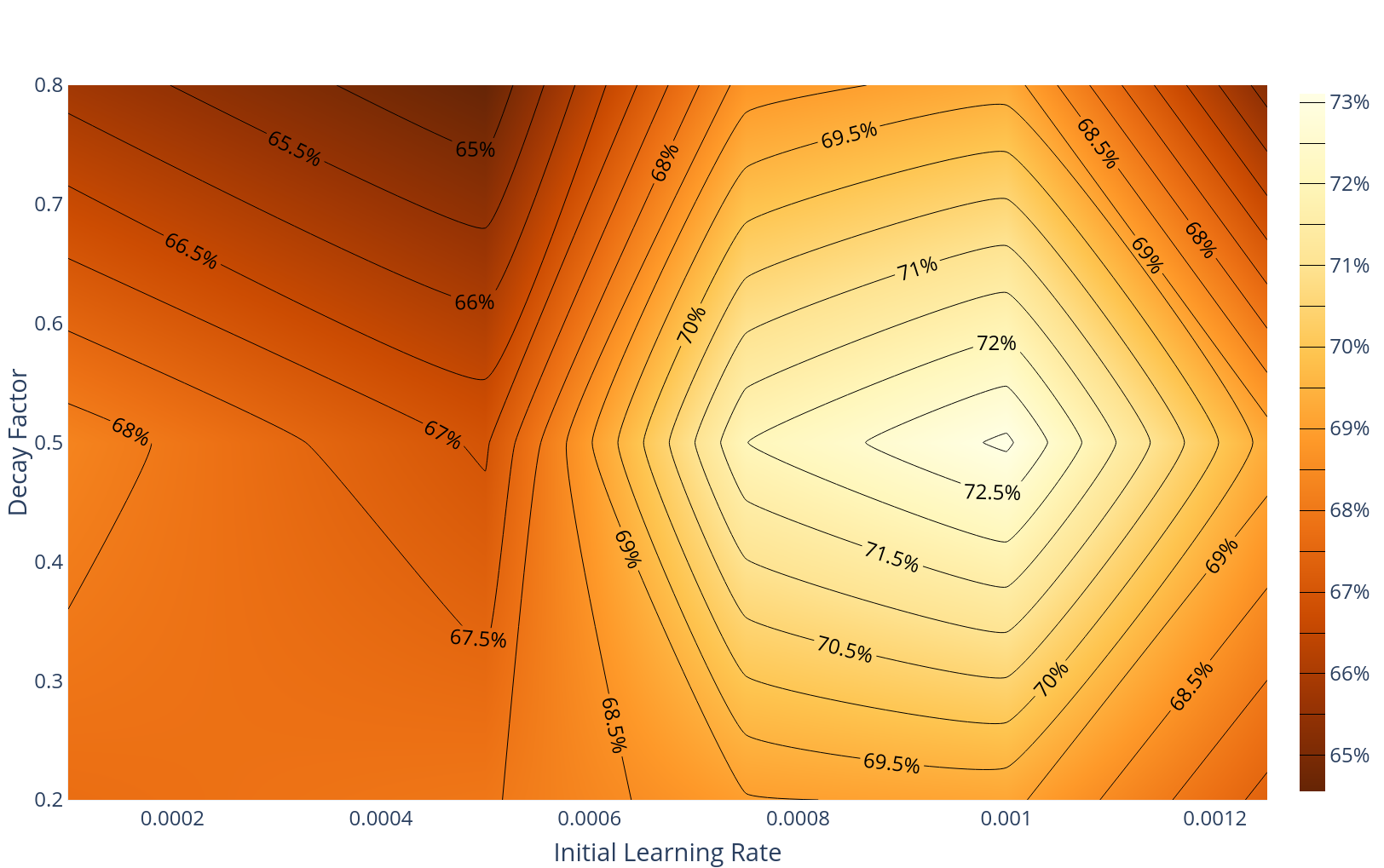}
\caption{Ablation over initial learning rate and decay factor. An optimum set of hyperparameters in downstream fine-tuning was found.}
\label{fig:ablatelr}
\end{figure*}

We perform the following ablations on the above structure on a pretrained UCF-101 model (downsampling of 2, color jitter + random grayscale, difference-based contrastive learning, no rotation; see the next section for details), fine-tuning on UCF-101 and evaluating on \emph{the UCF-101 test set}:
\begin{enumerate}
    \item We selected over the number of views, $K = (1, 2, 3, 4)$ (larger view counts hit memory limits), and found that more views translated to better performance (Pearson test two-tailed $p=0.07$).
    \item Lab dropout vs color jitter + random grayscale. We found that Lab dropout improved downstream performance ($p = 0.03$).
    \item We selected downsampling factors from $(1, 2, 3, 4)$ and found that less downsampling gave better results ($p = 0.0004$).
    \item We varied the initial learning rate along with the decay schedule. In this case, we decayed the learning rate every $30000$ steps at a fixed decay factor. Results are presented in Figure~\ref{fig:ablatelr}. 
\end{enumerate}
Based on this ablation study, we adopt a consistent fine-tuning procedure across all subsequent experiments: we use Lab dropout (instead of color jitter and random grayscale), no downsampling, $4$ views, an initial learning rate of $0.001$, and a learning rate decay factor of $0.5$.

\subsection{Ablation study on the effect of the self-supervised task layers and view generator on downstream performance}
We test VDIM, pretraining for $50000$ steps and fine-tuning for $20000$ steps on the UCF-$101$ dataset, ablating the augmentation across the following axes:
\begin{enumerate}
    \item Selecting fewer layers for the antecedent features, $\mathcal{J}$, comparing $\mathcal{J} = \{5, 6, 9\}$ to $\mathcal{J} = \{6, 9\}$ and $\mathcal{J} = \{9\}$. Removing layer $6$ had little effect, reducing performance by less than $1\%$. Removing both layers $5$ and $6$, however, greatly reduced performance, resulting in nearly a $10\%$ drop.
    \item Lab dropout vs random color jitter and grayscale. This had the strongest effect on downstream performance, with color jitter + random gray performing the best ($p=\num{9.5e-9}$).
    \item Randomly rotating the clip (spatially, the same rotation across the same view, maintaining the temporal order). No discernable effect of using rotation was found ($p=0.40$).
    \item With two sequences of frames, we select the second set to have a fixed offset of either $0$ (same start) or $32$ frames after the beginning of the first clip. We found that using no offset performed slightly better ($p=0.19$).
    \item Taking the temporal difference with consequent features (see Figure~\ref{fig:pretrain}). We found this improved performance ($p=0.05$).
    \item We downsampled the second clip by a factor of $1$ (no downsampling), $2$, or $3$. We found that when measured across all experimental parameters above, downsampling had a very small effect. However, when used in conjunction with difference features and color jitter, there is a small benefit to using downsampling ($p=0.12$). Increasing downsampling beyond $2$ had no significant effect on downstream performance.
\end{enumerate}
From these ablation experiments, the best model uses color jitter + random grayscale, no random rotation, and temporal differences for the consequent features from a subsequence with $0$ offset and a downsampling factor of $2$.

\begin{table}[ht]
\centering
\caption{Results on fine-tuning on UCF-$101$ after pretraining on UCF-$101$. Split 1 train / test was used in both pretraining and fine-tuning. Parameter counts are included for reference, when available. As we performed an ablation using the UCF-$101$ test set for model selection, some averages across the ablation are provided for reference.}
\begin{tabular}{@{}|l|l|l|@{}}
\hline
Model                                                      & Test acc (\%) & Params             \\
\hline
Shuffle and Learn~\citep{misra2016shuffle}                                          & 50.9          &                        \\
O3N~\citep{fernando2017self}                                                        & 60.3          &                        \\
DPC~\citep{han2019video}                                                        & 60.6          & 14.2M                  \\
Skip-Clip~\citep{el2019skip}                                                  & 64.4          & 14.2M                  \\
TCE~\citep{knights2020temporally}                                                        & 67.0          &                        \\
\hline
VDIM (avg ablate: color jitt) & \bf{69.0}         & \multirow{3}{*}{17.3M} \\
VDIM (avg ablate: color jitt, down, and diff feat) & 71.1         &                        \\
VDIM (ablation best)                                                & 74.1         &                        \\
\hline
\end{tabular}
\label{tab:ucf}
\end{table}

Our final comparison to several baselines pretrained and fine-tuned on UCF-101 is presented in Table~\ref{tab:ucf}. 
We compare against Shuffle and Learn~\citep{misra2016shuffle}, Dense Predictive Coding~\citep[DPC,][]{han2019video}, Temporally Coherent Embeddings~\citep[TCE,][]{knights2020temporally}, O3N~\citep{fernando2017self}, and Skip-Clip~\citep{el2019skip}.
Even if we consider the average performance of fine-tuned VDIM models pretrained under different sets of hyperparameters (across all ablation hyperparameters except using color-jitter and random grayscale, as this had the most obvious effect), VDIM outperforms the best baseline by $2\%$ on pretraining and fine-tuning on UCF-101. Our best result is the highest test accuracy known to us when training only on UCF-101 and without hand-crafted features~\citep[e.g., see][]{zou2012deep}.

\subsection{Kinetics-pretraining for downstream tasks}
We now show that pretraining VDIM on a large-scale dataset, Kinetics $600$~\cite{carreira2018short}, yields strong downstream performance on the UCF-101 and HMDB-$51$ datasets.
We pretrained on the Kinetics $600$ dataset for $100$k steps with a batch size of $240$ on $16$ GPUs with no rotation, no temporal difference\footnote{We omit temporal difference in this experiment for now because of unrelated hardware issues.}, downsampling of $2$, and color jitter + random grayscale.
For HMDB-$51$, the clips are generally much shorter than in UCF-$101$, resulting in 6-7 times fewer minibatches per epoch, so we adjusted the decay schedule from $3000$ steps to $400$ steps to accommodate\footnote{No hyperparameter tuning was performed on this decay schedule}; we also reduced the number of views to $2$.
We pretrain on the Kinetics dataset using the same procedure described in the ablation studies above, but for $30000$ training steps for UCF-$101$ and $4000$ steps for HBDB-$51$.
We fine-tune independently on all three UCF-$101$ or HMDB-$51$ splits and compute the final test accuracy from the average over test splits.

\begin{table}[ht]
\centering
\caption{A comparison of models pretrained on Kinetics-$600$ and fine-tuned on either UCF-101 or HMDB-51. Average test performance across all three splits is reported.}
\begin{tabular}{@{}|l|l|l|l|@{}}
\hline
Model     & UCF-101 acc & HMDB acc & Parameters \\
\hline
3D-RotNet~\citep{jing2018self} & 62.9        & 33.7     & 33.6M      \\
DPC~\citep{han2019video}       & 75.7        & 35.7     & 32.6M      \\
CBT~\citep{sun2019contrastive}       & {\bf 79.5}        & 44.5     & 20M        \\
VDIM      & {\bf 79.7}        &  {\bf 49.2}        & 17.3M      \\
\hline
\end{tabular}
\label{tab:kinetics}
\end{table}

We present results in Table~\ref{tab:kinetics}. Baselines include Contrastive Bidirectional Transformers~\citep[CBT,][]{sun2019contrastive}\footnote{The number of parameters for CBT was calculated from the stated 11M parameters for the visual transformer plus 9M parameters for S3D as stated in \citet{xie2017rethinking}.}, DPC, and 3D-RotNet~\citep{jing2018self}.
CBT uses curriculum learning during pretraining, first using rotation prediction~\citep{jing2018self}, followed by a contrastive loss on features processed by a transformer.
VDIM performs comparably to CBT on UCF-$101$ and outperforms on HMDB-$51$ by nearly $5\%$, despite the greater compute cost for training CBT.

\section{Conclusion and Acknowledgements}
We show that extracting and using feature-wise views as well as video-specific data augmentation in contrastive learning is an effective way to pretrain video models for downstream action recognition.
Specifically we show this can be done by extracting views from features naturally expressed in a deep spatio-temporal encoder, a method we call \emph{Video Deep InfoMax} (VDIM).
Our results show that there is strong promise in spatio-temporal encoders for unsupervised pretraining on video data, and that data augmentation on video data is an important component for unsupervised representation learning.
Finally, we would like to thank Tong Wang for his important contributions in developing this work.

\bibliographystyle{plainnat}
\bibliography{main}

\end{document}